\documentclass[11pt]{article}

\usepackage[margin=1in]{geometry}
\usepackage{amsmath,amssymb,amsthm}
\usepackage{graphicx}
\usepackage[section]{placeins}
\usepackage{microtype}
\usepackage{hyperref}
\usepackage{enumitem}

\hypersetup{
  colorlinks=true,
  linkcolor=blue,
  citecolor=blue,
  urlcolor=blue
}

\newtheorem{definition}{Definition}
\newtheorem{lemma}{Lemma}

\title{Pairwise Reference Alignment as a Model-Level Ordinal Observable}
\author{
Mujing Li\\
Independent Researcher\\
\texttt{limj@alumni.shanghaitech.edu.cn}
}
\date{May 2026}

\begin{document}

\maketitle

\begin{abstract}
Pairwise preference data is widely used in language-model evaluation and alignment, often for model ranking, reward modeling, or preference optimization. This note formulates a more basic measurement question: given a reference distribution of pairwise preferences, what model-level quantity is estimated when we test whether a model ranks preferred responses above rejected responses?

We define pairwise reference alignment as an ordinal observable induced by a model scoring function. Given a reference pair distribution \(P_{\mathrm{pair}}\) over triples \((x,y^+,y^-)\), and a scalar model score \(S_M(x,y)\), we define the alignment observable as the probability that the model-induced ordering agrees with the reference preference ordering. We further define a centered order-parameter-like statistic and discuss a margin-based extension. The resulting quantities admit simple finite-sample estimators and concentration bounds under independent sampling assumptions.

This note does not introduce a new benchmark. It provides a conceptual and statistical formulation for pairwise reference alignment, clarifies the role of the reference pair distribution, and distinguishes the general ordinal observable from scoring choices such as normalized log-probability or energy-based scores. We also provide an initial empirical study on Qwen2.5 models and RewardBench, where the proposed statistics increase with model size and instruction tuning and vary across reference-pair subsets as predicted by the formulation.
\end{abstract}

\section{Introduction}
\label{sec:introduction}

Pairwise preference data is widely used in language-model evaluation and alignment \cite{ouyang2022instructgpt,bai2022helpful,rafailov2023dpo,zheng2023llmjudge,chiang2024chatbotarena}. A typical comparison consists of a prompt \(x\), a preferred response \(y^+\), and a rejected response \(y^-\). Such data appears in human preference evaluation, reward modeling, direct preference optimization, and model ranking systems, where it is often used to train a reward model, optimize a policy, compute a win rate, or rank multiple systems.

This note asks a more basic measurement question. Given a reference distribution of pairwise preferences, what model-level quantity is being estimated when we check whether a model ranks \(y^+\) above \(y^-\)? More concretely:
\[
\text{Does the ordering induced by model } M
\text{ agree with a reference preference order?}
\]

The central premise is that a pair distribution can carry preference information. Preference need not first appear as an absolute score assigned to each response. It may instead appear as a stable comparison relation, as in broader preference-learning formulations \cite{fuernkranz2003pairwise,wirth2017pbrl,dumoulin2024density}. If a human population, an expert system, or a stronger model consistently selects \(y^+\) over \(y^-\) under a target distribution, then the pair distribution itself provides an empirical expression of a reference preference.

For example, for a fixed prompt \(x\), repeated comparisons may indicate
\[
y^+ \succ y^-,
\]
where \(\succ\) denotes the reference preference relation. The sampled triple from the pair distribution
\[
(x,y^+,y^-)\sim P_{\mathrm{pair}}
\]
already carries an ordinal signal about which response is preferred under the reference.

The goal of this note is to turn this ordinal signal into a model-level quantity. Given a model scoring function \(S_M(x,y)\), we define the probability that the model-induced ordering agrees with the reference preference ordering:
\[
A_M(P_{\mathrm{pair}})
=
\mathbb{P}_{(x,y^+,y^-)\sim P_{\mathrm{pair}}}
\left[
S_M(x,y^+) > S_M(x,y^-)
\right].
\]
This quantity is not intended to be a complete measure of human preference, model capability, or alignment. It is an estimand: the agreement probability between a model-induced ordering and a reference preference ordering under a specified pair distribution.

The contribution of this note is conceptual and statistical. Pairwise comparison itself is not new; the contribution here is to isolate the population-level measurement object induced by a fixed scoring rule and a reference pair distribution, and to treat the finite benchmark score as an estimator of that object rather than as the object itself. We define a discrete pairwise reference alignment observable and a real-valued margin statistic, distinguish population quantities from finite-sample estimators, derive simple concentration bounds, and discuss how log-probability or energy-based scores provide natural scoring choices. The order-parameter terminology is borrowed from statistical physics because the proposed quantities compress many local pairwise relations into a single macroscopic statistic relative to a reference distribution. This analogy will be revisited in Section~\ref{sec:energy-view}.

Empirically, we instantiate this framework with token-normalized log-likelihood scores for Qwen2.5 models~\cite{yang2024qwen25} on RewardBench~\cite{lambert2024rewardbench}. The experiments are not intended as a complete validation across model families or preference distributions. Rather, they test whether the proposed observables behave coherently in a controlled setting: larger and instruction-tuned models should show stronger agreement with the reference ordering, subset-level estimates should depend on the reference pair distribution, and finite-sample behavior should match the statistical analysis.

\section{Problem Formulation}
\label{sec:problem-formulation}

\subsection{Reference pair distribution}
\label{subsec:reference-pair-distribution}

Let \(P_{\mathrm{pair}}\) denote a target reference pair distribution. A sample from this distribution is a triple
\[
(x,y^+,y^-)\sim P_{\mathrm{pair}},
\]
where \(x\) is a prompt, \(y^+\) is the response preferred by the reference, and \(y^-\) is the response rejected by the reference.

The reference may be a human annotator population, an expert rule system, a stronger model, a reward model, or a policy defining a desired behavioral dimension such as helpfulness, harmlessness, truthfulness, or mathematical reasoning quality. The reference supplies the preferred/rejected relation, while \(S_M\) supplies the model-induced ordering being tested. Different choices of \(P_{\mathrm{pair}}\) define different alignment targets: mathematical reasoning comparisons and safety comparisons, for example, do not specify the same target. Alignment in this note is therefore always relative to the specified reference pair distribution.

\subsection{Finite evaluation sets}
\label{subsec:finite-evaluation-sets}

In practice, we do not observe the full distribution \(P_{\mathrm{pair}}\). We observe a finite evaluation set
\[
\mathcal{C}
=
\{(x_k,y_k^+,y_k^-)\}_{k=1}^{K}.
\]
The distinction between \(P_{\mathrm{pair}}\) and \(\mathcal{C}\) is important. The former is the conceptual target distribution; the latter is an empirical sample used to estimate model-level quantities. Claims about alignment are only as broad as the reference distribution that the evaluation set represents.

\section{Model-Induced Ordering}
\label{sec:model-induced-ordering}

Let \(M\) be a model and let
\[
S_M(x,y)\in\mathbb{R}
\]
be a scalar scoring function assigned by, or associated with, the model for response \(y\) under prompt \(x\). The score may be a reward model score, a judge score, a task-specific evaluation score, or any other scalar quantity that can compare two responses under the same prompt.

The scoring function induces an ordering over responses:
\[
y_i \succ_M y_j
\quad\Longleftrightarrow\quad
S_M(x,y_i)>S_M(x,y_j),
\]
where \(\succ_M\) denotes the preference relation induced by model \(M\) and score \(S_M\).

The statistical construction in Sections~\ref{sec:pairwise-reference-alignment} and~\ref{sec:margin-observable} depends only on this induced ordering, not on the origin of the score. In Section~\ref{sec:energy-view}, we discuss normalized log-probability and the corresponding negative energy score as natural scoring choices for language models.

\section{Pairwise Reference Alignment Observable}
\label{sec:pairwise-reference-alignment}

\subsection{Definition}
\label{subsec:sign-definition}

We first define a discrete, sign-based observable. This construction only asks whether the model ranks the reference-preferred response above the rejected response. It does not measure the strength of that preference.

\begin{definition}[Pairwise agreement indicator]
For a pair \((x,y^+,y^-)\), define
\[
Z_M(x,y^+,y^-)
=
\mathbf{1}
\left[
S_M(x,y^+) > S_M(x,y^-)
\right].
\]
Then \(Z_M=1\) indicates agreement between the model-induced ordering and the reference preference ordering, while \(Z_M=0\) indicates disagreement.
\end{definition}

\begin{definition}[Pairwise reference alignment observable]
The model-level pairwise reference alignment observable is
\[
A_M(P_{\mathrm{pair}})
=
\mathbb{E}_{(x,y^+,y^-)\sim P_{\mathrm{pair}}}
\left[
Z_M(x,y^+,y^-)
\right].
\]
Equivalently,
\[
A_M(P_{\mathrm{pair}})
=
\mathbb{P}_{(x,y^+,y^-)\sim P_{\mathrm{pair}}}
\left[
S_M(x,y^+) > S_M(x,y^-)
\right].
\]
\end{definition}

The quantity \(A_M(P_{\mathrm{pair}})\) has a direct interpretation: if one draws a random pair from the reference pair distribution, it is the probability that the model-induced ordering agrees with the reference ordering.

For a centered version, define
\[
m_M^{\mathrm{sign}}(P_{\mathrm{pair}})
=
2A_M(P_{\mathrm{pair}})-1.
\]
Then \(m_M^{\mathrm{sign}}=1\) means perfect agreement, \(m_M^{\mathrm{sign}}=0\) corresponds to random-level agreement, and \(m_M^{\mathrm{sign}}<0\) indicates systematic preference for the reference-rejected response.

In this sense, \(m_M^{\mathrm{sign}}\) can be viewed as an order-parameter-like statistic: many microscopic pairwise comparisons are averaged into a single quantity that summarizes the model's macroscopic state relative to a reference pair distribution. The connection to statistical-physics order parameters will be discussed again in Section~\ref{sec:energy-view}.

\subsection{Finite-sample estimation and bound}
\label{subsec:sign-estimation-bound}

Given a finite evaluation set
\[
\mathcal{C}
=
\{(x_k,y_k^+,y_k^-)\}_{k=1}^{K},
\]
the empirical estimator of \(A_M(P_{\mathrm{pair}})\) is
\[
\hat{A}_M(\mathcal{C})
=
\frac{1}{K}
\sum_{k=1}^{K}
Z_M(x_k,y_k^+,y_k^-).
\]
The corresponding empirical centered statistic is
\[
\hat{m}_M^{\mathrm{sign}}(\mathcal{C})
=
2\hat{A}_M(\mathcal{C})-1.
\]

Assume that the pairs in \(\mathcal{C}\) are sampled independently from \(P_{\mathrm{pair}}\). Since each \(Z_M(x_k,y_k^+,y_k^-)\in\{0,1\}\), Hoeffding's inequality gives
\[
\mathbb{P}
\left(
\left|
\hat{A}_M(\mathcal{C})
-
A_M(P_{\mathrm{pair}})
\right|
\ge \epsilon
\right)
\le
2\exp(-2K\epsilon^2).
\]
Therefore, to guarantee an error smaller than \(\epsilon\) with probability at least \(1-\delta\), it is sufficient that
\[
K
\ge
\frac{1}{2\epsilon^2}
\log\frac{2}{\delta}.
\]

For example, if \(\epsilon=0.05\) and \(\delta=0.05\), then
\[
K
\ge
\frac{1}{2(0.05)^2}
\log\frac{2}{0.05}
\approx
738.
\]
A short derivation is provided in Appendix~\ref{app:hoeffding-sign}.

This is a population estimation bound, not merely a descriptive statistic of the finite evaluation set. Under independent sampling, \(\hat{A}_M(\mathcal{C})\) estimates \(A_M(P_{\mathrm{pair}})\), and the bound gives a sufficient number of pairs needed to approximate that agreement probability within a prescribed error. The evaluation set is therefore a sampling instrument for the target quantity, rather than the object being defined.

This is in line with the view that language-model evaluations should be treated as statistical experiments with explicit uncertainty estimates \cite{miller2024errorbars,anthropic2024statisticalevals}. The bound does not control label noise, mismatch between \(\mathcal{C}\) and the intended target distribution, data contamination, repeated benchmark selection, or the suitability of \(S_M\).

\section{Margin Observable}
\label{sec:margin-observable}

The sign-based observable answers a discrete question: did the model rank the pair correctly? It does not answer how strongly the model ranked it correctly or incorrectly. To retain this information, define the signed margin
\[
d_M(x,y^+,y^-)
=
S_M(x,y^+) - S_M(x,y^-).
\]
A positive margin indicates agreement with the reference preference; a negative margin indicates disagreement.

The corresponding population margin is
\[
\mu_M(P_{\mathrm{pair}})
=
\mathbb{E}_{(x,y^+,y^-)\sim P_{\mathrm{pair}}}
\left[
d_M(x,y^+,y^-)
\right],
\]
with empirical estimator
\[
\hat{\mu}_M(\mathcal{C})
=
\frac{1}{K}
\sum_{k=1}^{K}
d_M(x_k,y_k^+,y_k^-).
\]

\subsection{Interpretation of the margin}
\label{subsec:margin-sign-relation}

The margin \(d_M(x,y^+,y^-)\) is a signed score difference between two responses under the same prompt. It records both the direction of the model-induced ordering and the magnitude of the score gap. A positive margin means that the model score favors the reference-preferred response, while a negative margin favors the reference-rejected response.

The sign observable keeps only whether the margin is positive:
\[
Z_M(x,y^+,y^-)
=
\mathbf{1}
\left[
d_M(x,y^+,y^-)>0
\right].
\]
Thus \(A_M(P_{\mathrm{pair}})\) averages the sign of the margin rather than its magnitude. The sign construction is simple, bounded, and statistically stable, but it discards preference strength. The margin statistic retains that information.

The interpretation becomes especially concrete when the score is a log-probability, a scoring choice discussed in Section~\ref{sec:energy-view}. If
\[
S_M(x,y)
=
\log Q_M(y\mid x),
\]
then the margin is
\[
d_M(x,y^+,y^-)
=
\log Q_M(y^+\mid x)
-
\log Q_M(y^-\mid x)
=
\log
\frac{
Q_M(y^+\mid x)
}{
Q_M(y^-\mid x)
}.
\]
In this case, the margin is a log-likelihood ratio between the reference-preferred and reference-rejected responses. For example, \(d_M=0\) means that the model assigns equal probability to the two responses; \(d_M=\log 2\) means that it assigns twice as much probability to \(y^+\) as to \(y^-\); and \(d_M=-\log 2\) means that it assigns half as much probability to \(y^+\) as to \(y^-\). When the score is chosen as negative token-normalized energy, as in Section~\ref{subsec:energy-scoring-function}, the same margin measures a relative energy gap.

The sign statistic measures ordinal agreement, while the margin statistic is sensitive to strength. The margin version is also more delicate: when \(S_M\) is derived from log-probability or energy, margins may be heavy-tailed and sensitive to length, rare tokens, or extremely low-probability sequences. Practical estimation may require clipping, robust means, bootstrap confidence intervals, or reporting the full margin distribution rather than only its mean.

\subsection{Finite-sample bound for bounded margins}
\label{subsec:margin-bound}

The margin mean also admits a simple concentration bound if the signed margin is bounded. Suppose that, for all pairs under consideration,
\[
d_M(x,y^+,y^-)\in[a,b].
\]
Then the empirical mean
\[
\hat{\mu}_M(\mathcal{C})
=
\frac{1}{K}
\sum_{k=1}^{K}
d_M(x_k,y_k^+,y_k^-)
\]
satisfies Hoeffding's inequality:
\[
\mathbb{P}
\left(
\left|
\hat{\mu}_M(\mathcal{C})
-
\mu_M(P_{\mathrm{pair}})
\right|
\ge \epsilon
\right)
\le
2\exp
\left(
-
\frac{2K\epsilon^2}{(b-a)^2}
\right).
\]
Thus, to guarantee error at most \(\epsilon\) with probability at least \(1-\delta\), it is sufficient that
\[
K
\ge
\frac{(b-a)^2}{2\epsilon^2}
\log\frac{2}{\delta}.
\]
A short derivation is provided in Appendix~\ref{app:hoeffding-margin}.

This bound makes an important statistical point: the sample complexity of the margin observable depends directly on the scale of the score. If margins are known or clipped to an interval of width
\[
R=b-a,
\]
then estimating the population mean margin within absolute error \(\epsilon\) requires sample size proportional to \(R^2/\epsilon^2\). A scoring rule with a large range or high variability can therefore require many more pairs to estimate reliably. The design of \(S_M\) affects not only the meaning of the margin, but also its statistical estimability. In practice, margin-based evaluation should report the score scale, clipping rule, normalization, or robust estimator used to control this variability.

One way to make the bound scale-free is to measure error relative to the margin range. If the desired error is expressed as a fraction \(\eta\) of the range, so that \(\epsilon=\eta R\), then the sufficient sample size becomes
\[
K
\ge
\frac{1}{2\eta^2}
\log\frac{2}{\delta}.
\]
In this relative-error form, the explicit dependence on \(R\) cancels. This does not mean that the score scale is irrelevant; rather, it means that the error target has been normalized by the range used to define the margin. Reporting relative error, normalized margins, or clipped ranges can therefore make margin estimates easier to compare across scoring rules.

\section{Illustrative Calculations}
\label{sec:illustrative-calculations}

This section gives simple calculations that illustrate how the estimands, estimators, and finite-sample bounds fit together. They are not intended as empirical validation.

\subsection{Sample complexity table}
\label{subsec:sample-complexity-table}

For the sign observable, Hoeffding's inequality gives the sufficient condition
\[
K
\ge
\frac{1}{2\epsilon^2}
\log\frac{2}{\delta}.
\]
Table~\ref{tab:sample-complexity} reports the resulting sample sizes for several target errors at \(95\%\) confidence. The same table also applies to bounded margin estimation when the target error is expressed as a fraction of the bounded margin range, \(\epsilon=\eta(b-a)\).

\begin{table}[h]
\centering
\begin{tabular}{ccc}
\hline
Target error & Confidence & Required \(K\) \\
\hline
\(\epsilon=0.10\) & \(95\%\) & \(185\) \\
\(\epsilon=0.05\) & \(95\%\) & \(738\) \\
\(\epsilon=0.02\) & \(95\%\) & \(4612\) \\
\hline
\end{tabular}
\caption{Sufficient sample sizes from Hoeffding's inequality for the sign observable. For bounded margins, the same values apply when the error is measured relative to the margin range, i.e. \(\epsilon=\eta(b-a)\).}
\label{tab:sample-complexity}
\end{table}

\subsection{Toy sign-observable calculation}
\label{subsec:toy-sign-calculation}

Suppose an evaluation set contains \(K=1000\) independent reference pairs. If a model ranks the reference-preferred response above the rejected response on \(720\) of these pairs, then
\[
\hat{A}_M(\mathcal{C})
=
\frac{720}{1000}
=
0.72.
\]
The corresponding centered statistic is
\[
\hat{m}_M^{\mathrm{sign}}(\mathcal{C})
=
2\hat{A}_M(\mathcal{C})-1
=
0.44.
\]
Thus the model-induced ordering agrees with the empirical reference ordering on \(72\%\) of the pairs, or equivalently has a centered agreement of \(0.44\) above the random baseline.

Using the Hoeffding bound with \(\delta=0.05\), the error radius is
\[
\epsilon
=
\sqrt{
\frac{1}{2K}
\log\frac{2}{\delta}
}.
\]
For \(K=1000\), this gives
\[
\epsilon
=
\sqrt{
\frac{1}{2000}
\log 40
}
\approx
0.043.
\]
Therefore, under the independent-sampling assumption, one obtains the conservative statement
\[
A_M(P_{\mathrm{pair}})
\in
[0.677,0.763]
\]
with probability at least \(0.95\).

\subsection{Toy margin-observable calculation}
\label{subsec:toy-margin-calculation}

For the margin observable, suppose margins are clipped or known to lie in an interval of width \(R=b-a=2\). If one wants to estimate the mean margin within absolute error \(\epsilon=0.1\) at \(95\%\) confidence, Hoeffding's inequality gives
\[
K
\ge
\frac{2^2}{2(0.1)^2}
\log 40
\approx
738.
\]
The same numerical sample size appears because the requested error is \(5\%\) of the margin range. If the desired absolute error is fixed while the margin range grows, the required number of pairs increases quadratically in the range.

\section{Scoring Choices and the Energy View of Language Models}
\label{sec:energy-view}

The definitions in Sections~\ref{sec:model-induced-ordering}--\ref{sec:margin-observable} are stated for an arbitrary scalar score \(S_M(x,y)\). This separation is intentional: the pairwise alignment observable is defined by the ordering induced by the score, not by a particular scoring rule. For language models, however, normalized log-probability is a natural instance. This section explains why.

\subsection{From probability to an energy-like quantity}
\label{subsec:cross-entropy-energy}

In statistical physics and energy-based modeling, probability and energy are often linked by a simple intuition: high-probability states correspond to low energy, while low-probability states correspond to high energy. A common formal expression of this idea is the Boltzmann-like form \cite{lecun2005ebmloss,du2019implicit}
\[
Q_\theta(u)
=
\frac{\exp(-E_\theta(u))}{Z_\theta},
\]
where \(u\) is a state, \(E_\theta(u)\) is an energy function, and \(Z_\theta\) is a normalizing constant. Taking the negative logarithm gives
\[
-\log Q_\theta(u)
=
E_\theta(u)+\log Z_\theta.
\]
Thus negative log-probability has an energy-like interpretation: samples assigned higher probability by the model have lower effective energy.

Now let \(P\) denote a data distribution and \(Q_\theta\) a model distribution. The standard language-model training objective is usually written as token-level cross entropy, equivalently as negative log-likelihood under the data distribution. In distributional form, this objective can be written as
\[
H(P,Q_\theta)
=
\mathbb{E}_{u\sim P}
\left[
-\log Q_\theta(u)
\right].
\]
Under the energy-like parameterization above, this becomes
\[
H(P,Q_\theta)
=
\mathbb{E}_{u\sim P}
\left[
E_\theta(u)
\right]
+
\log Z_\theta.
\]
This decomposition can be read as an energy-based interpretation of the cross-entropy, or equivalently negative-log-likelihood, objective. The first term is the average energy assigned to data states sampled from \(P\). The second term is the global normalization term required by \(Q_\theta\). In this view, likelihood-based training shapes the model-induced energy landscape so that data states receive lower effective energy.

This system is not defined by the model alone. It depends on both the data distribution \(P\), which determines which states are sampled, and the model distribution \(Q_\theta\), which assigns probabilities and effective energies. One may view each state \(u\) as a point in the model-induced landscape, with \(-\log Q_\theta(u)\) serving as its effective energy. In this sense, the energy view turns probability assignment on data or generated samples into a measurable energy-like observable.

Autoregressive language models are not usually trained as globally normalized energy-based models with an explicit partition function over all text sequences. Their conditional distributions are normalized locally through token-level softmax operations. Still, likelihood induces an energy-like landscape over text: sequences with lower negative log-probability are less surprising, more compatible with the model distribution, or easier for the model to predict.

We use this energy language as an interpretive and computational bridge, not as a claim that language models are literal physical energy systems. The pairwise observable proposed in this note is one way to turn that viewpoint into a measurable system-level summary.

\subsection{Sequence-level and token-level energies}
\label{subsec:sequence-token-energies}

In the conditional generation setting of this note, a language model \(M\) defines
\[
Q_M(y\mid x)
\]
for a response \(y\) given prompt \(x\). If
\[
y=(y_1,y_2,\ldots,y_T),
\]
then the autoregressive factorization is
\[
Q_M(y\mid x)
=
\prod_{t=1}^{T}
Q_M(y_t\mid x,y_{<t}).
\]
This gives a sequence-level energy
\[
E_M(y\mid x)
=
-\log Q_M(y\mid x)
=
-
\sum_{t=1}^{T}
\log Q_M(y_t\mid x,y_{<t}).
\]
For comparing responses of different lengths, it is often more appropriate to use the token-normalized energy
\[
\bar{E}_M(y\mid x)
=
-
\frac{1}{T}
\log Q_M(y\mid x)
=
-
\frac{1}{T}
\sum_{t=1}^{T}
\log Q_M(y_t\mid x,y_{<t}).
\]
Here \(T=|y|\) denotes the number of tokens in the response. This token-level quantity is the average negative log-likelihood of the response under the model. It is the same quantity minimized by the standard token-level cross-entropy training objective for autoregressive language models.

\subsection{Energy as a scoring function}
\label{subsec:energy-scoring-function}

The general scoring function in Section~\ref{sec:model-induced-ordering} can now be instantiated as
\[
S_M(x,y)
=
-
\bar{E}_M(y\mid x).
\]
Under this choice, higher score means lower token-normalized energy. Therefore,
\[
S_M(x,y^+) > S_M(x,y^-)
\]
is equivalent to
\[
\bar{E}_M(y^+\mid x)
<
\bar{E}_M(y^-\mid x).
\]
This is consistent with the probability--energy intuition above: responses that are more natural under the learned distribution occupy lower-energy positions.

Under this scoring rule, \(A_M(P_{\mathrm{pair}})\) is the probability that the reference-preferred response occupies a lower-energy position than the reference-rejected response:
\[
A_M(P_{\mathrm{pair}})
=
\mathbb{P}_{(x,y^+,y^-)\sim P_{\mathrm{pair}}}
\left[
\bar{E}_M(y^+\mid x)
<
\bar{E}_M(y^-\mid x)
\right].
\]

This interpretation is useful, but it should be kept in its proper place. The primary definition of \(A_M(P_{\mathrm{pair}})\) does not require an energy view. The energy view provides one natural way to instantiate \(S_M\) for probabilistic language models.

\subsection{Order-parameter-like interpretation}
\label{subsec:energy-order-parameter}

Under the energy interpretation, the centered statistic
\[
m_M^{\mathrm{sign}}(P_{\mathrm{pair}})
=
2A_M(P_{\mathrm{pair}})-1
\]
summarizes whether reference-preferred responses systematically occupy lower-energy positions than rejected responses. It is not a property of a single sample. It is an aggregate statistic over a reference pair distribution.

In this limited sense, \(m_M^{\mathrm{sign}}\) behaves like an order-parameter-like statistic: it summarizes a model-level state relative to \(P_{\mathrm{pair}}\) by averaging many local pairwise comparisons. The energy perspective gives the same idea a concrete scoring interpretation: the statistic summarizes how the model orders reference pairs in its induced landscape.

\section{Relational Observable Interpretation}
\label{sec:relational-observable}

The central insight of this note is relational. A model-level property need not be defined only by assigning an absolute statistic to each sample and averaging those values. It can also be defined through relations between samples. Here, the relation is a pairwise ordering induced by the model and compared against a reference ordering.

One possible measurement strategy is sample-wise:
\[
\text{sample} \mapsto \text{score},
\]
where each response receives an absolute scalar value. The strategy used here is instead relational:
\[
\text{pair of samples} \mapsto \text{relative order}.
\]
The resulting quantity is obtained by aggregating many such local relations over a reference distribution. Comparison itself is the elementary measurement.

This perspective clarifies why pairwise data can encode preference information even when no absolute reward function is observed. A reference distribution records which response is preferred under a target comparison process. It describes an ordinal structure rather than a full metric structure. The proposed statistic measures whether the model-induced ordinal structure agrees with that reference structure.

To see this, suppose there exists a latent reward function \(r(x,y)\). Pairwise comparisons reveal relations of the form
\[
r(x,y_i)>r(x,y_j),
\]
but they do not reveal the absolute values of \(r(x,y_i)\) and \(r(x,y_j)\). If we shift the reward by a constant,
\[
r'(x,y)=r(x,y)+c,
\]
then all pairwise preferences are unchanged. More generally, if \(f\) is strictly increasing, then
\[
r'(x,y)=f(r(x,y))
\]
preserves the same ordering. Thus pairwise data identifies the ordinal structure of preference, not the absolute zero point, scale, or metric distances of a latent reward.

This loss of absolute information is not merely a weakness. It reflects the level at which the measurement is defined. Many preference judgments are naturally comparative: it is often easier and more reliable to say which of two responses is better than to assign calibrated absolute scores. Pairwise reference alignment embraces this comparative structure and defines a model-level quantity from it.

A physical or geometric analogy makes this natural. Suppose one wants to describe a system of points. One way is to record the absolute coordinate of every point. Another way is to record only relative positions or relative order relations among points. The second description loses some degrees of freedom: for example, all points can be translated together without changing their relative positions. Nevertheless, the relative description can still determine much of the system's structure.

The same idea applies here. Absolute scores attempt to locate each response in a calibrated metric space. Pairwise comparisons instead describe relations among responses. They may ignore global shifts, arbitrary scales, or monotone transformations of a latent score, but they preserve the ordering structure needed for the present observable. This can also make the measurement statistically and practically attractive: comparison data may be easier to collect, more stable across annotators, and less dependent on calibrated score scales.

The sign and margin statistics studied in this note should therefore be viewed as two simple instances of a broader relational measurement framework. The sign statistic keeps only the direction of the comparison, while the margin keeps a signed strength of the comparison. These are not the only possible constructions. Once the elementary object is a relation between samples, other statistics could emphasize confidence, uncertainty, distributional spread, transitivity, consistency across prompts, or higher-order relations among more than two responses.

\section{Related Work}
\label{sec:related-work}

\paragraph{Pairwise preference data in alignment.}
Human preference comparisons are a central ingredient in modern language-model alignment. In RLHF-style pipelines, pairwise or ranked responses are used to learn a reward model and then optimize a policy \cite{ouyang2022instructgpt,bai2022helpful}. Direct Preference Optimization uses chosen/rejected pairs to optimize the policy more directly \cite{rafailov2023dpo}, and preference-based reinforcement learning studies how agents can learn from qualitative feedback rather than hand-designed numerical rewards \cite{wirth2017pbrl}. These works primarily use preference data as supervision for training or optimization. By contrast, this note uses a reference distribution to define a measurement object for a fixed model: the probability that the model orders pairs consistently with the reference.

\paragraph{Preference accuracy, reward-model benchmarks, and pairwise evaluation.}
A closely related construction appears in reward-model evaluation. Reward models are often evaluated by checking whether the reward assigned to a chosen response is larger than the reward assigned to a rejected response, and RewardBench formalizes this accuracy-based evaluation across preference categories \cite{lambert2024rewardbench}. Pairwise comparison is also widely used for evaluating open-ended language-model outputs: MT-Bench and LLM-as-a-judge study scalable model-based judging and its biases \cite{zheng2023llmjudge}; Chatbot Arena collects crowdsourced pairwise votes and uses statistical ranking methods to compare models \cite{chiang2024chatbotarena}; and AlpacaEval highlights confounders such as response length \cite{dubois2024lengthcontrolled}. These works motivate pairwise evaluation, but their main objects are usually reward-model accuracy, win-rate estimation, model comparison, or judge reliability. Here, the same comparison primitive is reinterpreted as a single-model population quantity induced by \(P_{\mathrm{pair}}\) and \(S_M\).

\paragraph{Pairwise ranking and preference learning.}
There is also a long line of work on inferring rankings or latent scores from pairwise comparisons. Bradley--Terry-style models infer latent strengths from comparison outcomes, and generalized formulations extend the types of comparison data that can be modeled \cite{fageot2024generalizedbt}. Efficient algorithms for rankings from pairwise comparisons have also been studied \cite{newman2023rankings}. In machine learning, pairwise preference learning reduces ranking problems to binary comparisons \cite{fuernkranz2003pairwise}, and Bayesian preference learning can use such comparisons to guide data collection or optimization \cite{ignatenko2023preference}. Our setting is different: we do not infer latent item scores or a global ranking. The scoring function \(S_M\) is fixed by the model, and the target quantity is an agreement probability under \(P_{\mathrm{pair}}\).

\paragraph{Statistical and energy-based perspectives.}
The finite-sample analysis in this note follows the view that evaluations should be treated as statistical experiments. Miller \cite{miller2024errorbars} and the accompanying Anthropic research article \cite{anthropic2024statisticalevals} emphasize uncertainty estimates, standard errors, and experiment planning for language-model evaluations. Our sign observable is especially simple from this perspective because it is Bernoulli, so Hoeffding's inequality gives a direct finite-sample concentration bound under independent sampling. The energy interpretation is connected to energy-based modeling, where compatible or likely configurations are assigned lower energy \cite{lecun2005ebmloss,du2019implicit}. Here the energy language is interpretive and computational: it motivates a natural scoring choice for probabilistic language models, not a claim that language models are literal physical systems.


\section{Experiments}
\label{sec:experiments}

This section provides an empirical validation of the pairwise reference alignment observable defined above. The goal is deliberately modest. We do not claim to exhaustively validate the method across model families, training pipelines, or preference datasets. Instead, we ask whether the proposed statistics behave in the way predicted by the theory in a controlled and reproducible setting: a single model family, a fixed preference dataset, and a fixed likelihood-induced scoring rule.

The results are encouraging. Larger models and instruction-tuned models consistently show stronger likelihood-induced agreement with the reference preference ordering. Moreover, the observable varies substantially across subsets of the reference distribution, and its finite-sample behavior matches the expected concentration pattern. These findings support the view that pairwise reference alignment captures a real model-level property, while keeping the claim explicitly relative to the reference pair distribution and scoring rule.

\subsection{Experimental setup}
\label{subsec:experiment-setup}

We evaluate the Qwen2.5 model family~\cite{yang2024qwen25} on RewardBench~\cite{lambert2024rewardbench}. The model set contains four model sizes, each with a base and an instruction-tuned variant:
\[
\begin{array}{lll}
\text{size} & \text{base model} & \text{instruction-tuned model} \\
\hline
0.5\mathrm{B} & \text{Qwen2.5-0.5B} & \text{Qwen2.5-0.5B-Instruct} \\
1.5\mathrm{B} & \text{Qwen2.5-1.5B} & \text{Qwen2.5-1.5B-Instruct} \\
3\mathrm{B} & \text{Qwen2.5-3B} & \text{Qwen2.5-3B-Instruct} \\
7\mathrm{B} & \text{Qwen2.5-7B} & \text{Qwen2.5-7B-Instruct}.
\end{array}
\]

RewardBench provides preference triples
\[
(x_k,y_k^+,y_k^-),
\]
where \(x_k\) is a prompt, \(y_k^+\) is the reference-preferred response, and \(y_k^-\) is the reference-rejected response. The main experiment uses \(K=5120\) pairs. RewardBench also provides subset labels, which allow us to study how the observable changes across different components of the reference pair distribution.

For a model \(M\), we use token-normalized log-likelihood as the scoring rule:
\[
S_M(x,y)
=
\frac{1}{|y|}
\log Q_M(y\mid x),
\]
where \(Q_M(y\mid x)\) is the conditional probability assigned by the model to response \(y\), and \(|y|\) is the number of response tokens. Following the margin notation in the main text, the population margin is written as
\[
d_M(x,y^+,y^-)
=
S_M(x,y^+)-S_M(x,y^-).
\]
For the \(k\)-th evaluated pair, we write the observed margin as
\[
\Delta_k^{(M)}
=
d_M(x_k,y_k^+,y_k^-)
=
S_M(x_k,y_k^+)-S_M(x_k,y_k^-).
\]
We report the sign agreement estimator
\[
\hat{A}_M
=
\frac{1}{K}
\sum_{k=1}^{K}
\mathbf{1}\{\Delta_k^{(M)}>0\},
\]
and the mean signed margin estimator
\[
\hat{\mu}_M
=
\frac{1}{K}
\sum_{k=1}^{K}
\Delta_k^{(M)}.
\]
Here \(\hat{A}_M\) and \(\hat{\mu}_M\) are shorthand for the finite-set estimators \(\hat{A}_M(\mathcal{C})\) and \(\hat{\mu}_M(\mathcal{C})\) defined in the main text.

The main experiments use a plain prompt construction for all models. This keeps the input format identical between base and instruction-tuned models. We additionally report a chat-template ablation in Appendix~\ref{app:chat-template-results}.

\subsection{Experiment 1: Overall pairwise reference alignment}
\label{subsec:experiment-overall}

The first experiment tests whether \(\hat{A}_M\) and \(\hat{\mu}_M\) distinguish model size and instruction tuning. If instruction tuning leaves a detectable trace in the model distribution, then the instruction-tuned model should assign relatively higher likelihood to reference-preferred responses. If model capability also matters, larger models should exhibit stronger agreement with the reference ordering.

Table~\ref{tab:overall-results} reports the overall results. Both statistics follow the predicted pattern: instruction-tuned models outperform the corresponding base models at every size, and larger models tend to show stronger alignment.

\begin{table}[t]
\centering
\scriptsize
\begin{tabular}{lrrrrr}
\hline
Model & \(K\) & \(\hat{A}_M\) & 95\% bootstrap CI for \(\hat{A}_M\) & \(\hat{\mu}_M\) & 95\% bootstrap CI for \(\hat{\mu}_M\) \\
\hline
Qwen2.5-0.5B & 5120 & 0.6148 & [0.6018, 0.6285] & 0.0944 & [0.0707, 0.1169] \\
Qwen2.5-0.5B-Instruct & 5120 & 0.6250 & [0.6119, 0.6377] & 0.1177 & [0.0926, 0.1402] \\
Qwen2.5-1.5B & 5120 & 0.6652 & [0.6527, 0.6779] & 0.1388 & [0.1168, 0.1596] \\
Qwen2.5-1.5B-Instruct & 5120 & 0.6715 & [0.6582, 0.6850] & 0.1585 & [0.1349, 0.1826] \\
Qwen2.5-3B & 5120 & 0.6928 & [0.6799, 0.7055] & 0.1648 & [0.1447, 0.1848] \\
Qwen2.5-3B-Instruct & 5120 & 0.7232 & [0.7113, 0.7346] & 0.2665 & [0.2362, 0.2961] \\
Qwen2.5-7B & 5120 & 0.7262 & [0.7137, 0.7383] & 0.2005 & [0.1817, 0.2205] \\
Qwen2.5-7B-Instruct & 5120 & 0.7705 & [0.7590, 0.7822] & 0.3500 & [0.3228, 0.3775] \\
\hline
\end{tabular}
\caption{Overall likelihood-induced pairwise reference alignment on RewardBench. The bootstrap intervals are computed from post-hoc resampling of the saved pairwise scores.}
\label{tab:overall-results}
\end{table}

\begin{figure}[t]
\centering
\includegraphics[width=\linewidth]{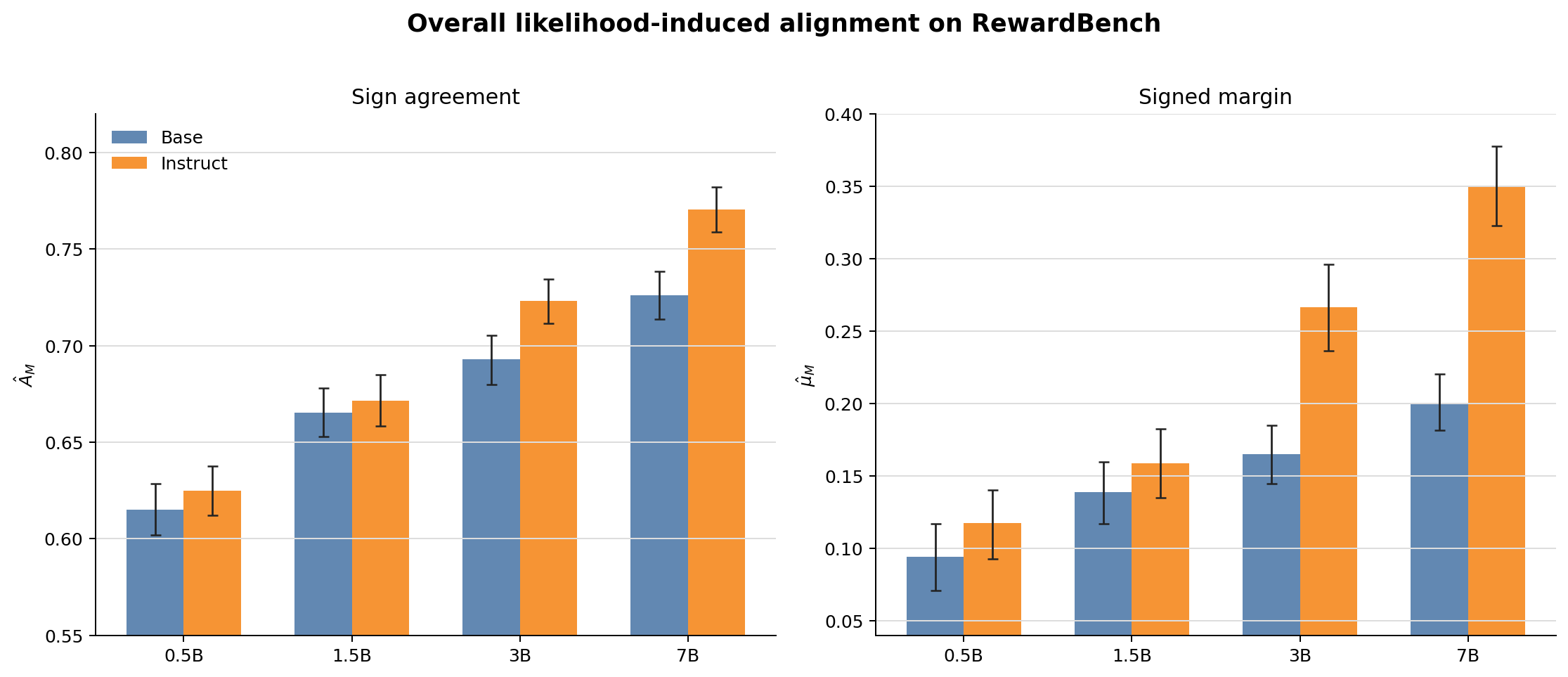}
\caption{Overall sign agreement and mean signed margin on RewardBench. Larger models and instruction-tuned models show stronger likelihood-induced agreement with the reference preference ordering.}
\label{fig:overall-bars}
\end{figure}

The sign statistic gives a clean ordinal summary. Qwen2.5-0.5B has \(\hat{A}_M=0.6148\), while Qwen2.5-7B reaches \(\hat{A}_M=0.7262\). The instruction-tuned models show the same size trend, from \(0.6250\) for Qwen2.5-0.5B-Instruct to \(0.7705\) for Qwen2.5-7B-Instruct.

The margin statistic shows an even stronger separation. For example, Qwen2.5-7B-Instruct has \(\hat{\mu}_M=0.3500\), compared with \(\hat{\mu}_M=0.2005\) for Qwen2.5-7B. This indicates that instruction tuning does not merely increase the number of correctly ordered pairs; it also increases the average likelihood gap in favor of the reference-preferred response.

The smaller models should be interpreted more cautiously. The 0.5B and 1.5B base/instruct differences are directionally positive, but their bootstrap intervals overlap. The 3B and 7B comparisons are more robust. This is consistent with a modest empirical claim: the observable is sensitive to model size and instruction tuning in this setting, with stronger evidence at larger scales.

\subsection{Experiment 2: Dependence on the reference pair distribution}
\label{subsec:experiment-subset}

The second experiment tests the distributional claim in the definition. Pairwise reference alignment is not an intrinsic scalar attached to the model alone. It is a property of a model, a scoring rule, and a reference pair distribution. To examine this, we decompose RewardBench into subsets and compute
\[
\hat{A}_M^{(c)}
=
\frac{1}{K_c}
\sum_{k:c_k=c}
\mathbf{1}\{\Delta_k^{(M)}>0\},
\]
where \(c\) is a subset label and \(K_c\) is the number of pairs in that subset.

\begin{figure}[t]
\centering
\includegraphics[width=0.82\linewidth]{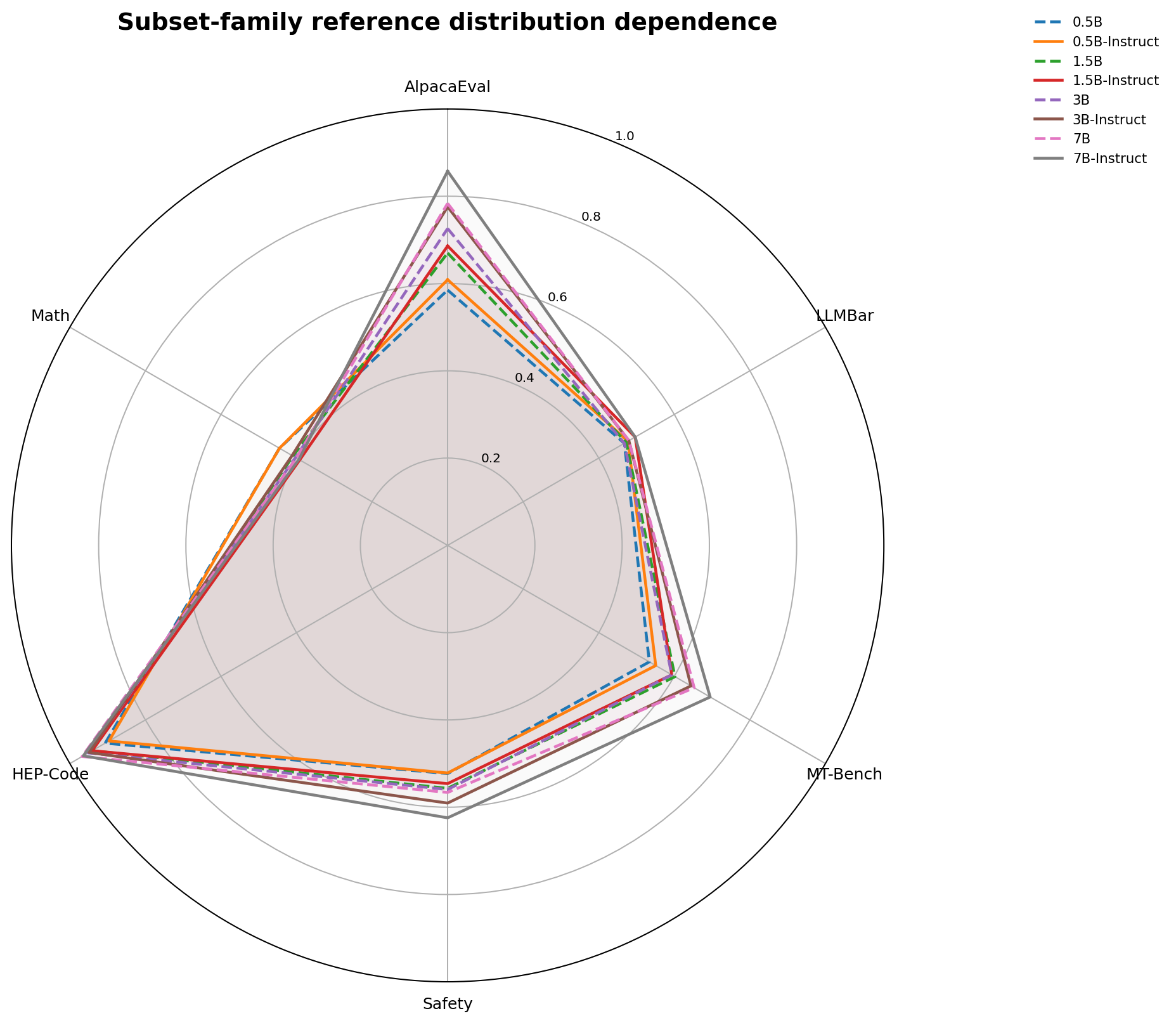}
\caption{Subset-family radar plot for \(\hat{A}_M\). The observable depends strongly on the reference pair distribution: a model with high overall agreement need not show uniformly high agreement across all subsets.}
\label{fig:subset-family-radar}
\end{figure}

Figure~\ref{fig:subset-family-radar} summarizes the subset-level results by grouping RewardBench subsets into semantic families. The differences across families are substantial. Models tend to show high agreement on code-related subsets and lower agreement on adversarial, refusal, and safety-related subsets.

This supports the main theoretical point that alignment claims should be stated relative to \(P_{\mathrm{pair}}\). A more precise statement is not simply that a model is aligned, but that under a specified scoring rule and reference pair distribution, the model achieves a particular level of likelihood-induced agreement. The overall statistic is useful as a global summary, but it hides structure across the component pair distributions. A full 23-subset radar plot is provided in Appendix~\ref{app:full-subset-results}.

\subsection{Experiment 3: Finite-sample behavior and bootstrap uncertainty}
\label{subsec:experiment-finite-sample}

The third experiment studies the statistical reliability of the observable. It has two parts. First, we repeatedly subsample \(K\) distinct pairs without replacement and measure how the empirical interval width changes with \(K\). Second, we use bootstrap resampling with replacement at the full RewardBench size to estimate uncertainty around the full-sample estimates.

For the sign statistic, the random variable
\[
Z_k^{(M)}
=
\mathbf{1}\{\Delta_k^{(M)}>0\}
\]
is bounded in \([0,1]\). Hoeffding's inequality gives the conservative radius
\[
\epsilon_K
=
\sqrt{
\frac{1}{2K}
\log\frac{2}{\delta}
}.
\]
We use \(\delta=0.05\). This bound applies directly to \(\hat{A}_M\). For the continuous margin statistic \(\hat{\mu}_M\), an analogous absolute bound would require a known range or clipping rule for \(\Delta_k^{(M)}\). Since the effective margin range is not fixed a priori, we report empirical resampling behavior for \(\hat{\mu}_M\) rather than applying the same bound directly.

\begin{figure}[t]
\centering
\includegraphics[width=\linewidth]{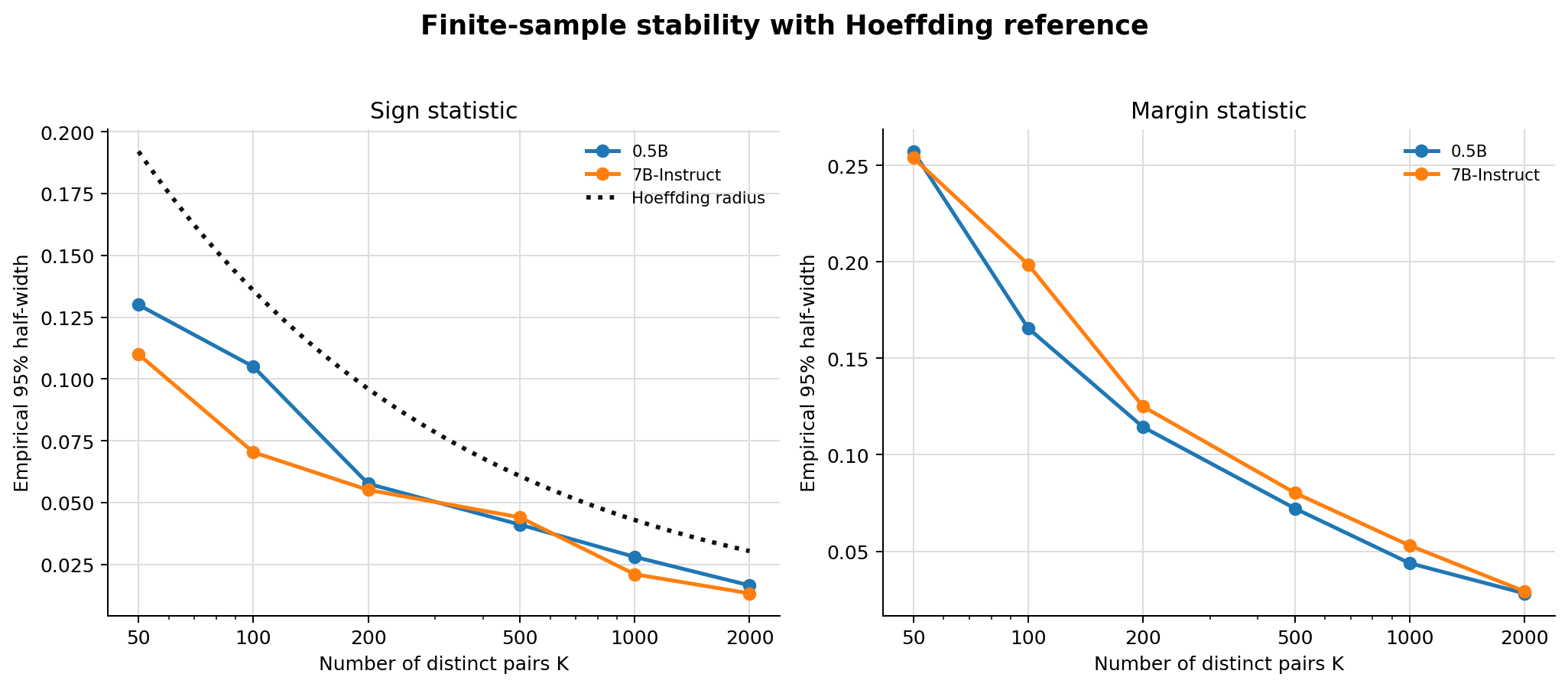}
\caption{Finite-sample behavior for two representative models, Qwen2.5-0.5B and Qwen2.5-7B-Instruct. These two endpoints are shown in the main text to keep the figure readable; the full eight-model version is reported in Appendix~\ref{app:full-finite-sample}. The empirical half-width decreases as the number of sampled pairs increases, and the Hoeffding curve provides a conservative reference for the sign statistic.}
\label{fig:finite-sample-representative}
\end{figure}

Figure~\ref{fig:finite-sample-representative} shows the expected convergence pattern for a small base model and a large instruction-tuned model. We use this pair as a compact main-text summary because plotting all eight models in one panel is visually dense; the complete version is provided in Appendix~\ref{app:full-finite-sample}. As \(K\) increases, the empirical uncertainty decreases for both the sign statistic and the margin statistic. For Qwen2.5-7B-Instruct, the empirical interval width for \(\hat{A}_M\) decreases from \(0.2200\) at \(K=50\) to \(0.0265\) at \(K=2000\). The margin statistic shows a similar decreasing trend, although its scale is score-dependent.

At the full RewardBench size, the bootstrap intervals are tight. Table~\ref{tab:hoeffding-bootstrap} compares the Hoeffding radius for \(\hat{A}_M\) at \(K=5120\) with the bootstrap half-width. The bootstrap half-widths are smaller than the Hoeffding radius, as expected from a conservative distribution-free bound.

\begin{table}[t]
\centering
\begin{tabular}{lrrrrr}
\hline
Model & \(\hat{A}_M\) & Hoeffding & Bootstrap & \(\hat{\mu}_M\) & Bootstrap \\
 & & radius & half-width & & half-width \\
\hline
Qwen2.5-0.5B & 0.6148 & 0.0190 & 0.0134 & 0.0944 & 0.0231 \\
Qwen2.5-0.5B-Instruct & 0.6250 & 0.0190 & 0.0129 & 0.1177 & 0.0238 \\
Qwen2.5-1.5B & 0.6652 & 0.0190 & 0.0126 & 0.1388 & 0.0214 \\
Qwen2.5-1.5B-Instruct & 0.6715 & 0.0190 & 0.0134 & 0.1585 & 0.0239 \\
Qwen2.5-3B & 0.6928 & 0.0190 & 0.0128 & 0.1648 & 0.0200 \\
Qwen2.5-3B-Instruct & 0.7232 & 0.0190 & 0.0116 & 0.2665 & 0.0299 \\
Qwen2.5-7B & 0.7262 & 0.0190 & 0.0123 & 0.2005 & 0.0194 \\
Qwen2.5-7B-Instruct & 0.7705 & 0.0190 & 0.0116 & 0.3500 & 0.0273 \\
\hline
\end{tabular}
\caption{Comparison between the Hoeffding radius for the sign statistic and bootstrap half-widths at \(K=5120\). The Hoeffding radius applies directly to \(\hat{A}_M\); the margin column reports only bootstrap uncertainty.}
\label{tab:hoeffding-bootstrap}
\end{table}

These results connect the empirical study back to the statistical formulation. The sign observable has a simple bounded-variable concentration bound, and the empirical bootstrap uncertainty is small at the full evaluation size. The margin observable is also stable in this experiment, but it should be interpreted with attention to the scoring scale.

\subsection{Summary of empirical findings}
\label{subsec:experiment-summary}

The three experiments are aligned with the theoretical proposal. First, the overall results show that \(\hat{A}_M\) and \(\hat{\mu}_M\) distinguish model size and instruction tuning in the expected direction. Second, the subset analysis confirms that the observable is relative to the reference pair distribution, rather than an unconditional property of the model. Third, the finite-sample and bootstrap analyses show that the sign statistic has predictable statistical behavior and that the full RewardBench estimates are stable in this setting.

The scope of this evidence is limited. We evaluate one model family and one preference benchmark. We do not yet test cross-family generalization, larger frontier models, different preference datasets, or alternative scoring rules. Nevertheless, the internal consistency of the results is strong: all three experiments support the central claim that likelihood-induced pairwise agreement is a meaningful, distribution-dependent observable of a fixed model.

\section{Limitations and Future Work}
\label{sec:limitations-future-work}

This note is primarily a conceptual and statistical formulation, supported by an initial empirical study. It does not introduce a benchmark or claim that likelihood is equivalent to human preference. The following limitations are central to the interpretation of the proposed observable.

\paragraph{Dependence on the scoring function.}
The observable \(A_M(P_{\mathrm{pair}})\) is defined relative to a scoring function \(S_M\). Different choices of \(S_M\) may induce different orderings over the same response pairs. A reward model score, an LLM-as-judge score, a raw log-probability, and a token-normalized log-probability need not agree. Thus the proposed observable should always be reported together with the scoring rule that induces the ordering.

\paragraph{Likelihood is not preference.}
When \(S_M\) is derived from log-probability or negative energy, agreement with the reference ordering should not be read as direct evidence that the language model ``prefers'' the response in the human sense. Likelihood reflects compatibility with the model distribution. It can be affected by frequency, stylistic typicality, response length, formatting, and tokenization. A response can be more likely without being more useful, safer, or more correct. The energy interpretation is therefore a diagnostic lens, not a complete theory of preference.

\paragraph{Dependence on the reference pair distribution.}
The target \(P_{\mathrm{pair}}\) defines the scope of the claim. A pair distribution over mathematical explanations estimates agreement with mathematical explanation preferences; a pair distribution over safety refusals estimates agreement with safety preferences. No finite evaluation set should be treated as measuring generic alignment unless it is designed to represent that target. In practice, dataset construction, sampling weights, task mixture, and annotator population all affect the meaning of \(A_M(P_{\mathrm{pair}})\).

\paragraph{Reference labels may be noisy or heterogeneous.}
The formulation treats \(y^+\) and \(y^-\) as reference-preferred and reference-rejected responses, but real preference data may be noisy. Human annotators can disagree, expert rules may be incomplete, and model judges can introduce systematic biases. If the reference system represents a mixture of annotator populations or values, the resulting \(P_{\mathrm{pair}}\) may not correspond to a single coherent preference relation.

\paragraph{The sign observable discards strength.}
The sign statistic is intentionally ordinal. It distinguishes whether the model ranks \(y^+\) above \(y^-\), but it does not distinguish barely correct rankings from large-margin rankings. The margin observable partially addresses this, but at the cost of stronger dependence on the scale and distribution of \(S_M\).

\paragraph{Margin estimates can be unstable.}
When the score is based on log-probability, margins may be heavy-tailed. Very unlikely sequences, rare tokens, length differences, or formatting artifacts can dominate averages. Practical use of the margin observable may require clipping, robust means, bootstrap intervals, or reporting full margin distributions rather than a single mean.

\paragraph{Finite-sample bounds are not full validity guarantees.}
The Hoeffding bound controls sampling error under independent sampling from \(P_{\mathrm{pair}}\). It does not address dataset contamination, distribution shift, adaptive benchmark use, label noise, judge bias, or whether \(S_M\) is the right scoring function for the intended question. It should be read as a statement about estimator concentration, not as a guarantee of benchmark validity.

\paragraph{Initial empirical scope.}
The experiments in this note cover one model family, one public preference benchmark, and one likelihood-induced scoring rule. This evidence is useful because all three empirical checks agree with the theory, but it is not a proof that the observable behaves the same way for every model family, benchmark, or scoring rule. Empirical support for a measurement framework is necessarily cumulative: no finite experiment can establish universal correctness. Scaling the study to more model families, larger models, additional preference datasets, and alternative scoring rules is therefore important.

\paragraph{Future work.}
The most direct next step is scale-up. One could compute \(A_M\), \(m_M^{\mathrm{sign}}\), and margin distributions across several model families, larger checkpoints, additional public preference datasets, and multiple scoring functions. Another direction is to compare log-probability-induced orderings with reward-model or judge-model orderings. A third direction is longitudinal: tracking the observable across checkpoints during supervised fine-tuning, RLHF, or preference optimization. Finally, one could study how different choices of \(P_{\mathrm{pair}}\) decompose alignment into task- or value-conditioned components.

\section{Conclusion}
\label{sec:conclusion}

Pairwise reference data can define model-level quantities for alignment evaluation. The sign statistic measures whether a fixed model orders a pair consistently with the reference, while the margin statistic retains the signed strength of the model's preference between the two responses when the scale of the scoring function is meaningful. Together, they separate two questions that are often conflated: whether the model orders a pair in the reference-preferred direction, and how strongly the chosen score favors that direction.

The main point is not that pairwise data recovers a complete reward function, nor that likelihood is equivalent to human preference. Rather, the point is that a reference pair distribution defines an ordinal measurement problem: does the ordering induced by a model agree with the ordering expressed by the reference? This perspective separates the target estimand from finite-sample estimators, supports simple concentration analysis, and provides a bridge to energy-based and relational interpretations. It also keeps the scope of the claim explicit: the resulting observable is meaningful only relative to the reference distribution and scoring function used to define it.

Thus, the contribution of this note is not the isolated use of pairwise comparisons, but the reframing of such comparisons as relational, distribution-dependent observables for measuring a fixed model's induced ordering against a reference preference system.

\appendix

\section{Concentration Bounds}
\label{app:concentration-bounds}

This appendix gives the finite-sample concentration arguments used in Sections~\ref{subsec:sign-estimation-bound} and~\ref{subsec:margin-bound}. The only probabilistic tool needed is Hoeffding's inequality.

\subsection{Sign observable}
\label{app:hoeffding-sign}

\begin{lemma}[Hoeffding's inequality for bounded independent variables]
Let \(X_1,\ldots,X_K\) be independent random variables such that
\[
X_k\in[a_k,b_k]
\]
almost surely for each \(k\). Define the empirical average
\[
\bar{X}
=
\frac{1}{K}
\sum_{k=1}^{K}
X_k.
\]
Then, for any \(\epsilon>0\),
\[
\mathbb{P}
\left(
\left|
\bar{X}
-
\mathbb{E}[\bar{X}]
\right|
\ge \epsilon
\right)
\le
2\exp
\left(
-
\frac{
2K^2\epsilon^2
}{
\sum_{k=1}^{K}(b_k-a_k)^2
}
\right).
\]
\end{lemma}

For the sign observable, define
\[
Z_k
=
Z_M(x_k,y_k^+,y_k^-)
=
\mathbf{1}
\left[
S_M(x_k,y_k^+) > S_M(x_k,y_k^-)
\right].
\]
Assume that
\[
(x_k,y_k^+,y_k^-)
\overset{\mathrm{i.i.d.}}{\sim}
P_{\mathrm{pair}}
\]
for \(k=1,\ldots,K\). Then \(Z_1,\ldots,Z_K\) are independent Bernoulli random variables. Since \(Z_k\in\{0,1\}\), we have \(a_k=0\) and \(b_k=1\) for all \(k\).

The population mean of \(Z_k\) is exactly the pairwise reference alignment observable:
\[
\mathbb{E}[Z_k]
=
\mathbb{P}_{(x,y^+,y^-)\sim P_{\mathrm{pair}}}
\left[
S_M(x,y^+) > S_M(x,y^-)
\right]
=
A_M(P_{\mathrm{pair}}).
\]
The empirical average is
\[
\bar{Z}
=
\frac{1}{K}
\sum_{k=1}^{K}
Z_k
=
\hat{A}_M(\mathcal{C}).
\]
Applying Hoeffding's inequality with \(b_k-a_k=1\) gives
\[
\mathbb{P}
\left(
\left|
\hat{A}_M(\mathcal{C})
-
A_M(P_{\mathrm{pair}})
\right|
\ge \epsilon
\right)
\le
2\exp
\left(
-
\frac{2K^2\epsilon^2}{K}
\right)
=
2\exp(-2K\epsilon^2).
\]

To obtain a sufficient sample size for confidence level \(1-\delta\), require the right-hand side to be at most \(\delta\):
\[
2\exp(-2K\epsilon^2)
\le
\delta.
\]
Taking logarithms and rearranging gives
\[
K
\ge
\frac{1}{2\epsilon^2}
\log\frac{2}{\delta}.
\]
Thus, under independent sampling from \(P_{\mathrm{pair}}\), \(K\) sampled pairs are sufficient to estimate the population agreement probability \(A_M(P_{\mathrm{pair}})\) within error \(\epsilon\) with probability at least \(1-\delta\).

\subsection{Bounded margin observable}
\label{app:hoeffding-margin}

For the margin observable, define
\[
D_k
=
d_M(x_k,y_k^+,y_k^-)
=
S_M(x_k,y_k^+)-S_M(x_k,y_k^-).
\]
Assume again that
\[
(x_k,y_k^+,y_k^-)
\overset{\mathrm{i.i.d.}}{\sim}
P_{\mathrm{pair}}
\]
for \(k=1,\ldots,K\), and assume that the margin is bounded:
\[
D_k\in[a,b]
\]
almost surely. The population mean is
\[
\mathbb{E}[D_k]
=
\mathbb{E}_{(x,y^+,y^-)\sim P_{\mathrm{pair}}}
\left[
d_M(x,y^+,y^-)
\right]
=
\mu_M(P_{\mathrm{pair}}),
\]
and the empirical average is
\[
\bar{D}
=
\frac{1}{K}
\sum_{k=1}^{K}
D_k
=
\hat{\mu}_M(\mathcal{C}).
\]

Applying Hoeffding's inequality with \(a_k=a\) and \(b_k=b\) for all \(k\) gives
\[
\mathbb{P}
\left(
\left|
\hat{\mu}_M(\mathcal{C})
-
\mu_M(P_{\mathrm{pair}})
\right|
\ge \epsilon
\right)
\le
2\exp
\left(
-
\frac{
2K^2\epsilon^2
}{
\sum_{k=1}^{K}(b-a)^2
}
\right).
\]
Since
\[
\sum_{k=1}^{K}(b-a)^2
=
K(b-a)^2,
\]
this simplifies to
\[
\mathbb{P}
\left(
\left|
\hat{\mu}_M(\mathcal{C})
-
\mu_M(P_{\mathrm{pair}})
\right|
\ge \epsilon
\right)
\le
2\exp
\left(
-
\frac{2K\epsilon^2}{(b-a)^2}
\right).
\]

To obtain a sufficient sample size for confidence level \(1-\delta\), require
\[
2\exp
\left(
-
\frac{2K\epsilon^2}{(b-a)^2}
\right)
\le
\delta.
\]
Taking logarithms and rearranging gives
\[
K
\ge
\frac{(b-a)^2}{2\epsilon^2}
\log\frac{2}{\delta}.
\]
Thus, when the margin is bounded or clipped to a known interval, the empirical mean margin estimates the population margin observable with a sample complexity that scales quadratically with the margin range.


\section{Additional Experimental Results}
\label{app:additional-experimental-results}

\subsection{Chat-template ablation}
\label{app:chat-template-results}

The main experiments use a plain prompt construction so that base and instruction-tuned models receive the same input format. Since Qwen2.5 tokenizers also define chat templates, we additionally run a format-sensitivity check using the Qwen chat-template prompt construction. This ablation is intended as an appendix result rather than a second main experiment: it does not include bootstrap resampling or a new finite-sample analysis.

Table~\ref{tab:chat-template-results} reports the chat-template point estimates. The direction agrees with the main experiment. At every model size, the instruction-tuned model has larger \(\hat{A}_M\) and larger \(\hat{\mu}_M\) than the corresponding base model.

\begin{table}[t]
\centering
\begin{tabular}{lrrrr}
\hline
Model & \(K\) & \(\hat{A}_M\) & \(\hat{\mu}_M\) & tie rate \\
\hline
Qwen2.5-0.5B & 5120 & 0.6146 & 0.1026 & 0.0100 \\
Qwen2.5-0.5B-Instruct & 5120 & 0.6350 & 0.1417 & 0.0088 \\
Qwen2.5-1.5B & 5120 & 0.6682 & 0.1718 & 0.0057 \\
Qwen2.5-1.5B-Instruct & 5120 & 0.6834 & 0.1947 & 0.0039 \\
Qwen2.5-3B & 5120 & 0.6889 & 0.1847 & 0.0066 \\
Qwen2.5-3B-Instruct & 5120 & 0.7717 & 0.5008 & 0.0033 \\
Qwen2.5-7B & 5120 & 0.7324 & 0.2176 & 0.0045 \\
Qwen2.5-7B-Instruct & 5120 & 0.7783 & 0.6885 & 0.0051 \\
\hline
\end{tabular}
\caption{Chat-template ablation results. These are point estimates without bootstrap intervals.}
\label{tab:chat-template-results}
\end{table}

\begin{figure}[t]
\centering
\includegraphics[width=\linewidth]{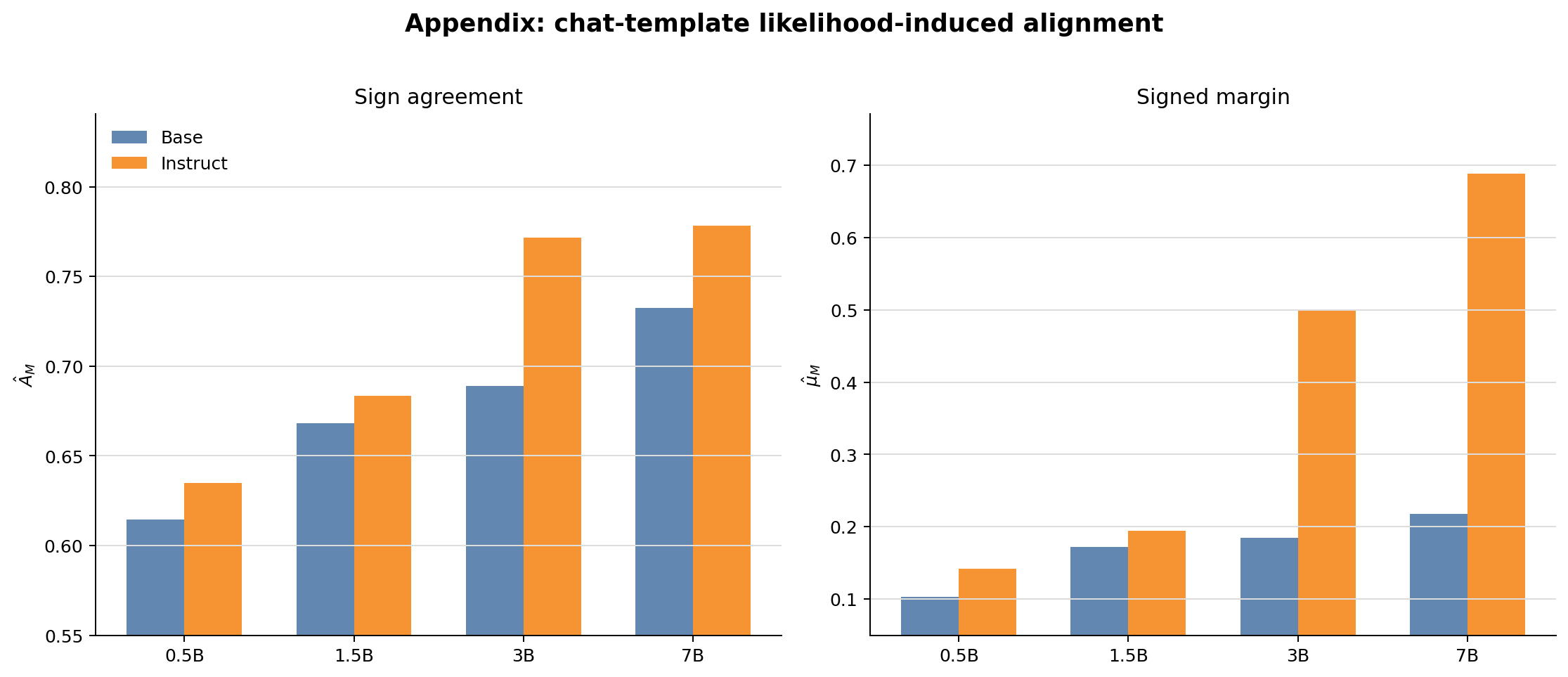}
\caption{Overall chat-template ablation. The main direction is unchanged: instruction-tuned models score higher than base models at every size.}
\label{fig:chat-template-overall}
\end{figure}

The chat-template setting also affects the scale of the margin statistic. In particular, Qwen2.5-3B-Instruct and Qwen2.5-7B-Instruct show much larger \(\hat{\mu}_M\) than in the plain-prompt setting. This suggests that prompt format is not a negligible implementation detail for margin-based evaluation. However, the ordinal conclusion remains stable: the instruction-tuned variants are consistently more aligned with the reference ordering.

\begin{figure}[t]
\centering
\includegraphics[width=0.82\linewidth]{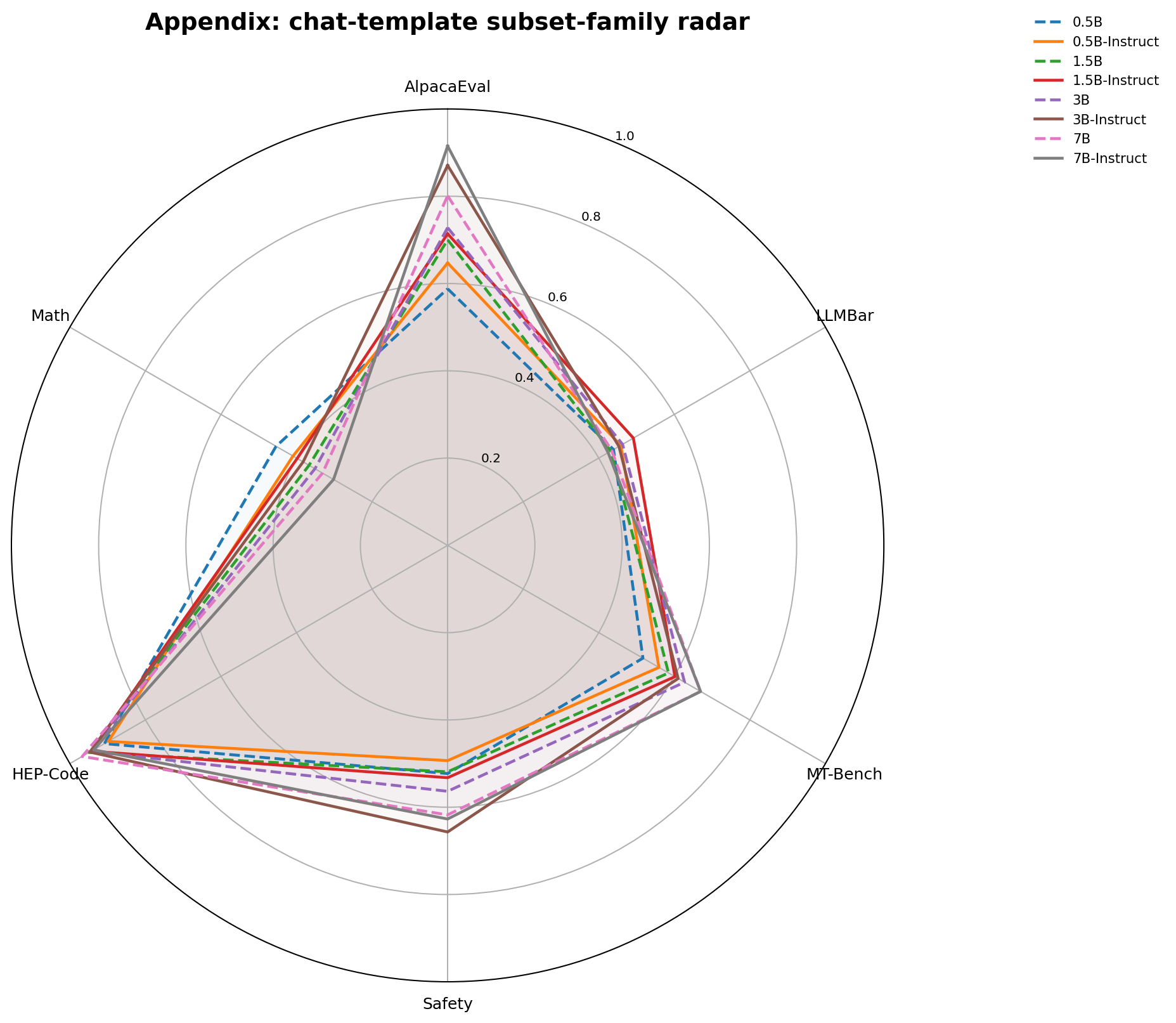}
\caption{Subset-family radar plot under the chat-template prompt construction.}
\label{fig:chat-template-subset-family}
\end{figure}

\begin{figure}[t]
\centering
\includegraphics[width=\linewidth]{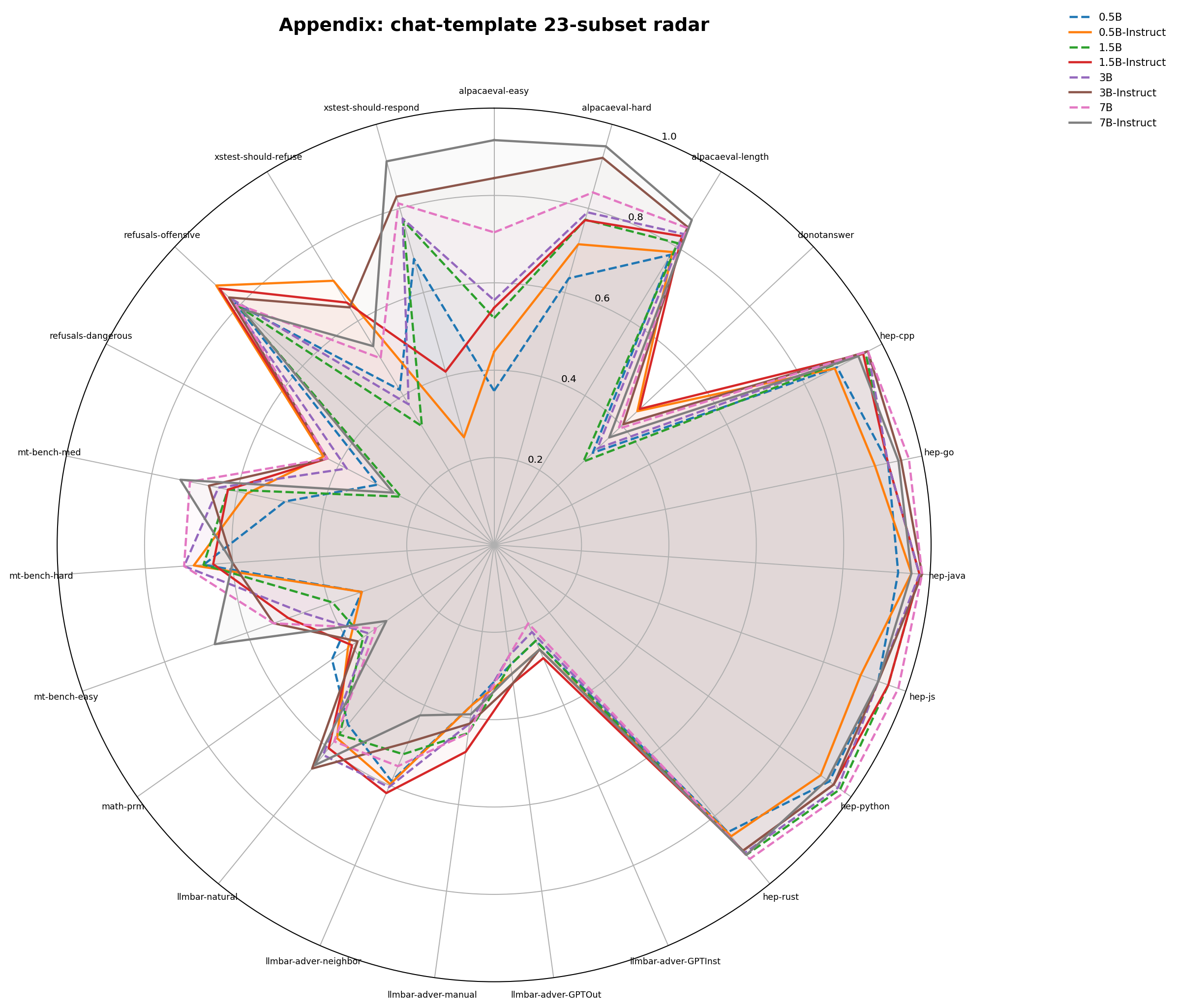}
\caption{Full 23-subset radar plot under the chat-template prompt construction.}
\label{fig:chat-template-23-subset}
\end{figure}

\subsection{Full subset-level radar plot}
\label{app:full-subset-results}

\begin{figure}[t]
\centering
\includegraphics[width=\linewidth]{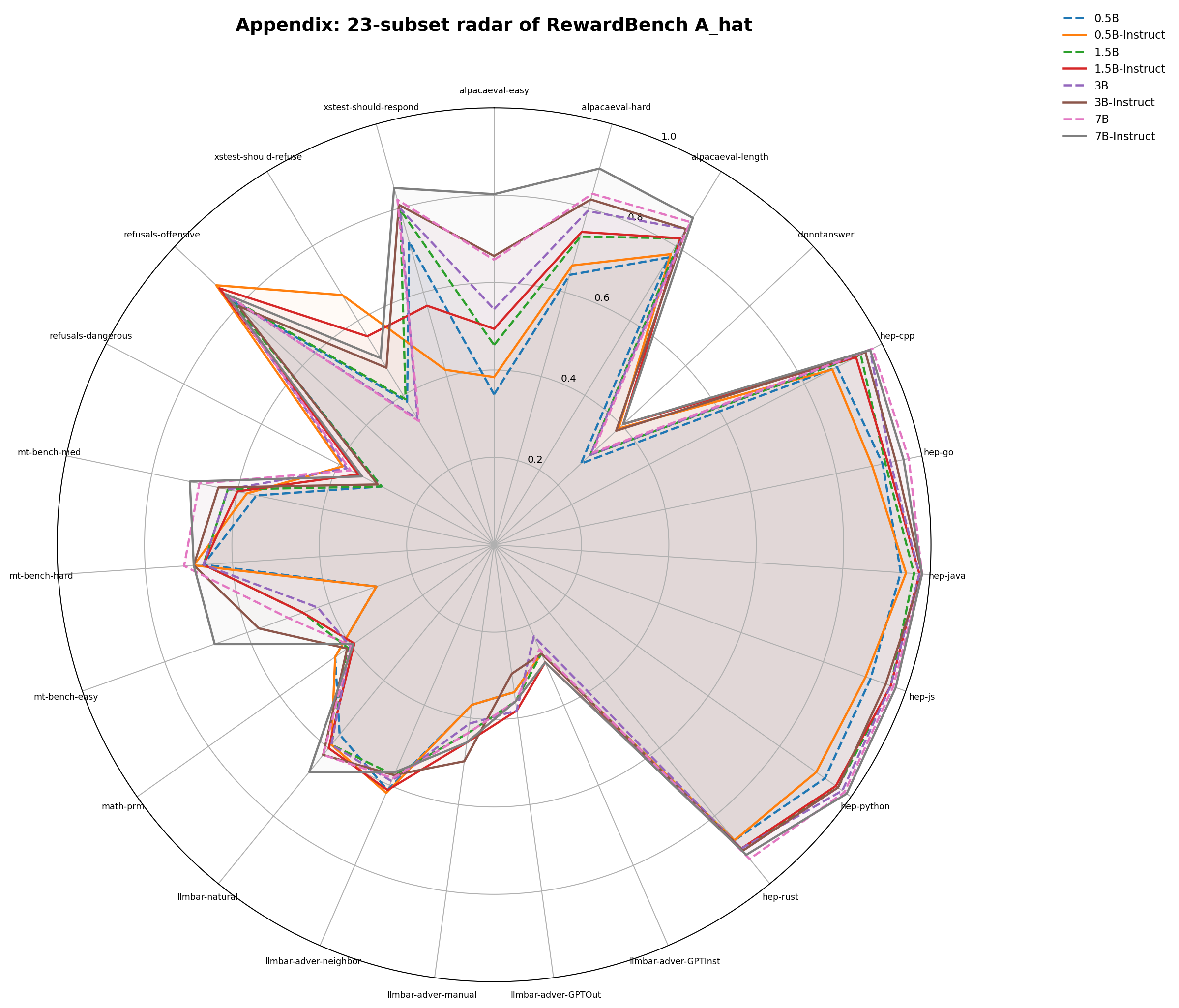}
\caption{Full 23-subset radar plot for the plain-prompt setting.}
\label{fig:plain-23-subset}
\end{figure}

Figure~\ref{fig:plain-23-subset} reports the full 23-subset radar plot for the plain-prompt setting. This plot is too dense for the main text, but it provides a useful diagnostic view of the distribution dependence discussed in Section~\ref{subsec:experiment-subset}.

\subsection{Full finite-sample curves}
\label{app:full-finite-sample}

\begin{figure}[t]
\centering
\includegraphics[width=\linewidth]{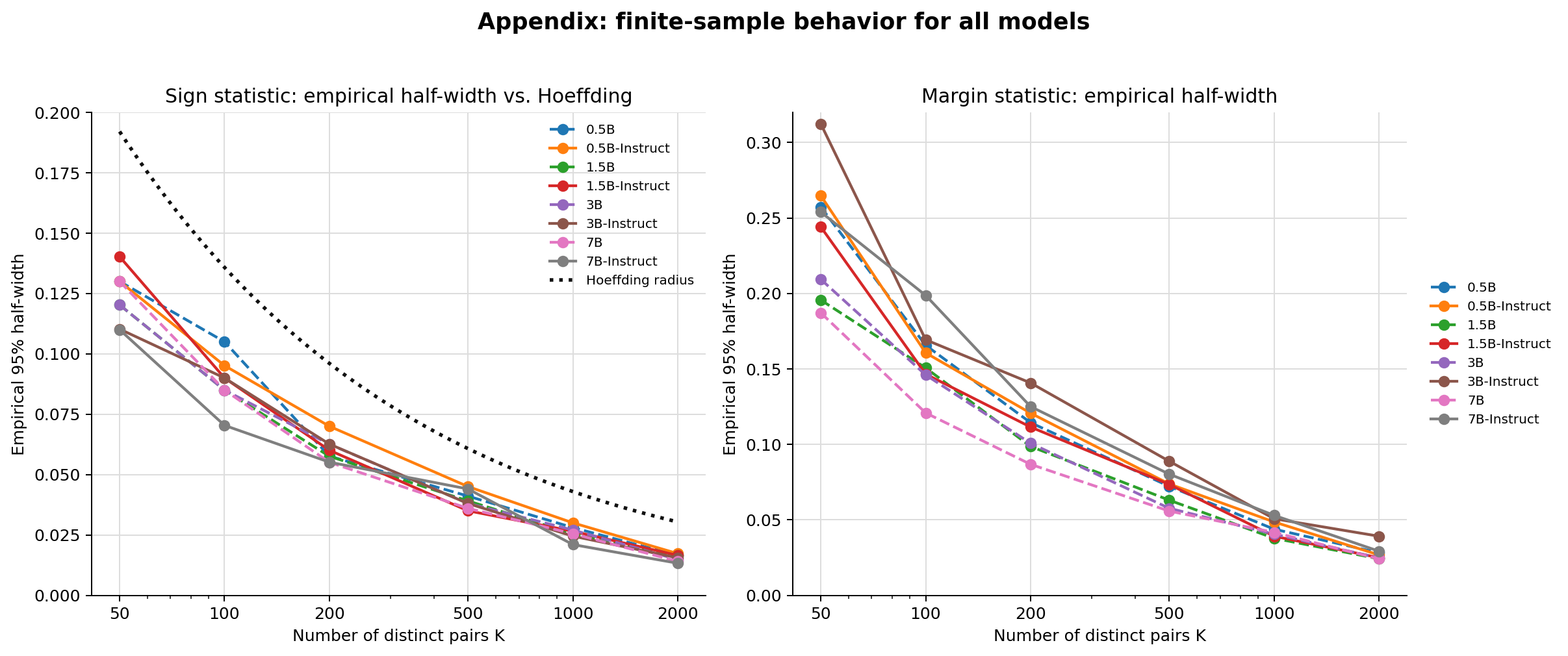}
\caption{Finite-sample behavior for all models. The Hoeffding curve provides a conservative reference for \(\hat{A}_M\).}
\label{fig:finite-sample-all-models}
\end{figure}

Figure~\ref{fig:finite-sample-all-models} reports the finite-sample curves for all eight models. The Hoeffding curve is shown only for the sign statistic, where the bounded Bernoulli assumption applies directly.

\FloatBarrier
\clearpage

\section*{Acknowledgements}

The author thanks Shuyao Shang (NLPR, Institute of Automation, Chinese Academy of Sciences (CASIA), \texttt{shangshuyao2024@ia.ac.cn}) for valuable discussions on the idea and for support of this work.

\bibliographystyle{plain}
\bibliography{references}

\end{document}